\documentclass{article}

\usepackage{microtype}
\usepackage{graphicx}
\usepackage{subcaption}
\usepackage{booktabs} 

\usepackage{hyperref}

\usepackage[preprint]{icml2026}

\usepackage{amsmath}
\usepackage{amssymb}
\usepackage{mathtools}
\usepackage{amsthm}
\usepackage{listings}

\usepackage[capitalize,noabbrev]{cleveref}

\theoremstyle{plain}
\newtheorem{theorem}{Theorem}[section]
\newtheorem{proposition}[theorem]{Proposition}
\newtheorem{lemma}[theorem]{Lemma}

\theoremstyle{definition}
\newtheorem{definition}[theorem]{Definition}
\newtheorem{assumption}[theorem]{Assumption}
\theoremstyle{remark}

\usepackage[textsize=tiny]{todonotes}

\usepackage{bm}
\usepackage{bbm}

\newcommand{\<}{\langle}
\renewcommand{\>}{\rangle}

\newcommand{\R}{\mathbbm{R}}
\newcommand{\C}{\mathbbm{C}}
\renewcommand{\L}{\mathcal{L}}
\newcommand{\M}{\mathcal{M}}

\newcommand{\mtx}{\bm} 
\newcommand{\vct}{\bm} 

\icmltitlerunning{A Bifurcation Theory Framework for Gradient Descent on the Edge of Stability}

\begin{document}

\twocolumn[
  \icmltitle{A Bifurcation Theory Framework for Gradient Descent on the Edge of Stability}



  \icmlsetsymbol{equal}{*}

  \begin{icmlauthorlist}
    \icmlauthor{Eric Gan}{indp}
  \end{icmlauthorlist}

  \icmlaffiliation{indp}{Independent Researcher}

  \icmlcorrespondingauthor{Eric Gan}{egan8@ucla.edu}

  \icmlkeywords{Machine Learning, ICML, Edge of Stability}

  \vskip 0.3in
]



\printAffiliationsAndNotice{}  

\begin{abstract}
The Edge of Stability (EoS) phenomenon, where gradient descent operates with sharpness exceeding the classical convergence threshold yet the loss decreases over long timescales, is ubiquitous in modern deep learning but remains poorly understood in realistic settings. Prior rigorous analyses have been largely confined to scalar or low-dimensional losses with specific structural forms. In this work, we develop a bifurcation theory framework for gradient descent on the edge of stability that applies directly to overparameterized neural networks. By decomposing the training dynamics into components normal and tangent to the manifold of minimizers, we show that stable EoS training arises from a flip bifurcation in the normal direction, governed by the sign of the first Lyapunov coefficient, while the tangent dynamics drift toward regions of decreasing sharpness. Under mild spectral and geometric assumptions on the loss landscape, we prove convergence to the minimizing manifold when training at the EoS threshold. As a corollary, we recover and unify prior results: we show that the product-stability condition of \citet{gan2026product} is an instance of our framework.
\end{abstract}

\section{Introduction} \label{sec:intro}

A central puzzle in modern machine learning optimization is the Edge of Stability (EoS) phenomenon, documented empirically by \citet{cohen2022gradient}. In standard gradient descent analysis, convergence is guaranteed when the learning rate $\eta$ and sharpness $\lambda$, defined as the largest eigenvalue of the loss Hessian, satisfy the relation $\eta \lambda < 2$. Yet in practice, models routinely train with $\eta \lambda > 2$: the sharpness rises above this classical threshold, the loss evolves non-monotonically, and yet convergence reliably occurs. This behavior cannot be explained by classical arguments, motivating a richer theoretical framework.

A productive line of work has studied EoS in minimalist settings \cite{zhu2023understanding, wang2023good, song2023trajectory, gan2026product}. These works offer clean, rigorous analyses of EoS dynamics, but they share a fundamental limitation: they are restricted to low-dimensional, specially structured losses that do not capture the geometry of modern neural network training.

A key feature of realistic neural network loss landscapes is the presence of a manifold of minimizers. Overparameterized networks have far more parameters than training examples, and as a result the loss landscape contains a continuous set of global minima \cite{draxler2019essentiallybarriersneuralnetwork, simsek2021geometrylosslandscapeoverparameterized}. The Hessian at any minimum is therefore rank-deficient, with a nontrivial kernel aligned with the tangent space of the minimizing manifold. 

In this work, we develop a framework that analyzes EoS dynamics around such a minimizing manifold $\M$. Our approach rests on a decoupling of the normal and tangent dynamics with respect to $\M$. In the normal direction, the dynamics undergo a flip bifurcation, the classical period-doubling bifurcation of discrete dynamical systems, whose stability is governed by the first Lyapunov coefficient $c_1$. When $c_1 > 0$, stable period-2 oscillations exist in the normal direction for learning rates slightly above the EoS threshold, preventing divergence. In the tangent direction, the two-step dynamics drift along $\M$ in the direction of decreasing sharpness. Together, these two effects stabilize the training dynamics and allow the iterates to converge.

We further show in Section 5 that the product-stability framework of \citet{gan2026product} emerges naturally as a special case. For losses of the form $\L(x, y) = f(xy)$, we prove that the first Lyapunov coefficient $c_1$ of $\L$ is controlled by the product-stability of $f$. This establishes a direct bridge between the concrete scalar analysis of prior work and the general bifurcation framework developed here.

Our main contributions are as follows:

\begin{itemize}
    \item We develop a bifurcation theory framework for gradient descent at the edge of stability that accommodates the manifold-of-minimizers geometry of overparameterized networks.

    \item We decompose the EoS dynamics into normal and tangent components, deriving explicit dynamical equations for each and showing how they cooperate to produce convergence. We prove convergence to the minimizing manifold at the EoS threshold under mild conditions, with the first Lyapunov coefficient as the key stability criterion.

    \item We show that the product-stability condition of Gan (2026) is a special case of our framework, unifying prior minimalist analyses with the general bifurcation framework.
\end{itemize}

\section{Related Work}

\textbf{Edge of Stability.} \citet{cohen2022gradient} provided the original systematic empirical documentation of EoS training dynamics, observing that the sharpness of deep networks rises to $\frac{2}{\eta}$ and then oscillates around this threshold while the loss decreases in the long run. Subsequent work has pursued theoretical explanations from several directions. \citet{damian2023selfstabilization} argue that third-order terms in the Taylor expansion of the loss contribute to sharpness self-stabilization. \citet{ma2022quadratic} analyze a multiscale loss structure with a subquadratic property that prevents divergence.

\textbf{Minimalist EoS analyses.} A productive approach has been to study small, analytically tractable models that nonetheless capture EoS behavior. \citet{wang2022large, liang2025gradient} prove EoS convergence for the quadratic loss $(xy - a)^2$. \citet{ahn2023learning} prove convergence for losses $l(xy)$ satisfying a convexity and subquadratic assumption. \citet{song2023trajectory} take a bifurcation-theoretic perspective on trajectory alignment for the same class of functions. \citet{zhu2023understanding} analyze the multilayer squared loss $(x^2 y^2 - 1)^2$ and establish sharpness adaptivity - the phenomenon that the limiting sharpness is close to but slightly below $\frac{2}{\eta}$. \citet{wang2023good} study EoS convergence for a class of functions based on a notion of degree of regularity. \citet{gan2026product} unifies all of these results under the product-stability condition and extends them to binary cross entropy; our framework subsumes that work as a special case.

\textbf{Bifurcation theory in optimization.} \citet{chen2023edge} first used the two-step fixed point structure to analyze GD beyond the EoS threshold, implicitly invoking bifurcation-theoretic ideas. \citet{mulayoff2026stability} explicitly apply bifurcation theory to derive conditions for stable period-2 oscillations in the multivariate setting, proving existence of stable periodic orbits when the first Lyapunov coefficient is positive. Our work builds directly on their framework and extends it to handle the additional structure present in the EoS setting: the full manifold of minimizers, the drift dynamics along that manifold, and the resulting convergence.

\textbf{Manifolds of minimizers.} The geometry of loss landscapes in overparameterized networks has been studied extensively. \citet{draxler2019essentiallybarriersneuralnetwork} and others have documented the existence of connected loss-zero manifolds. \citet{simsek2021geometrylosslandscapeoverparameterized} analyze symmetry-induced degeneracies. \citet{li2022happenssgdreacheszero} develop a mathematical framework for SGD dynamics along manifolds of minimizers. \citet{arora2022understanding} analyze EoS dynamics along manifolds of minimizers, but require specialized loss functions that do not match practical settings. The novelty of our contribution relative to these works is the use of bifurcation theory to define the role of the first Lyapunov coefficient in governing stability during training at the EoS threshold.
\section{Bifurcation Theory}

\begin{figure}[tbp] 
  \centering
  \includegraphics[width=0.5\textwidth]{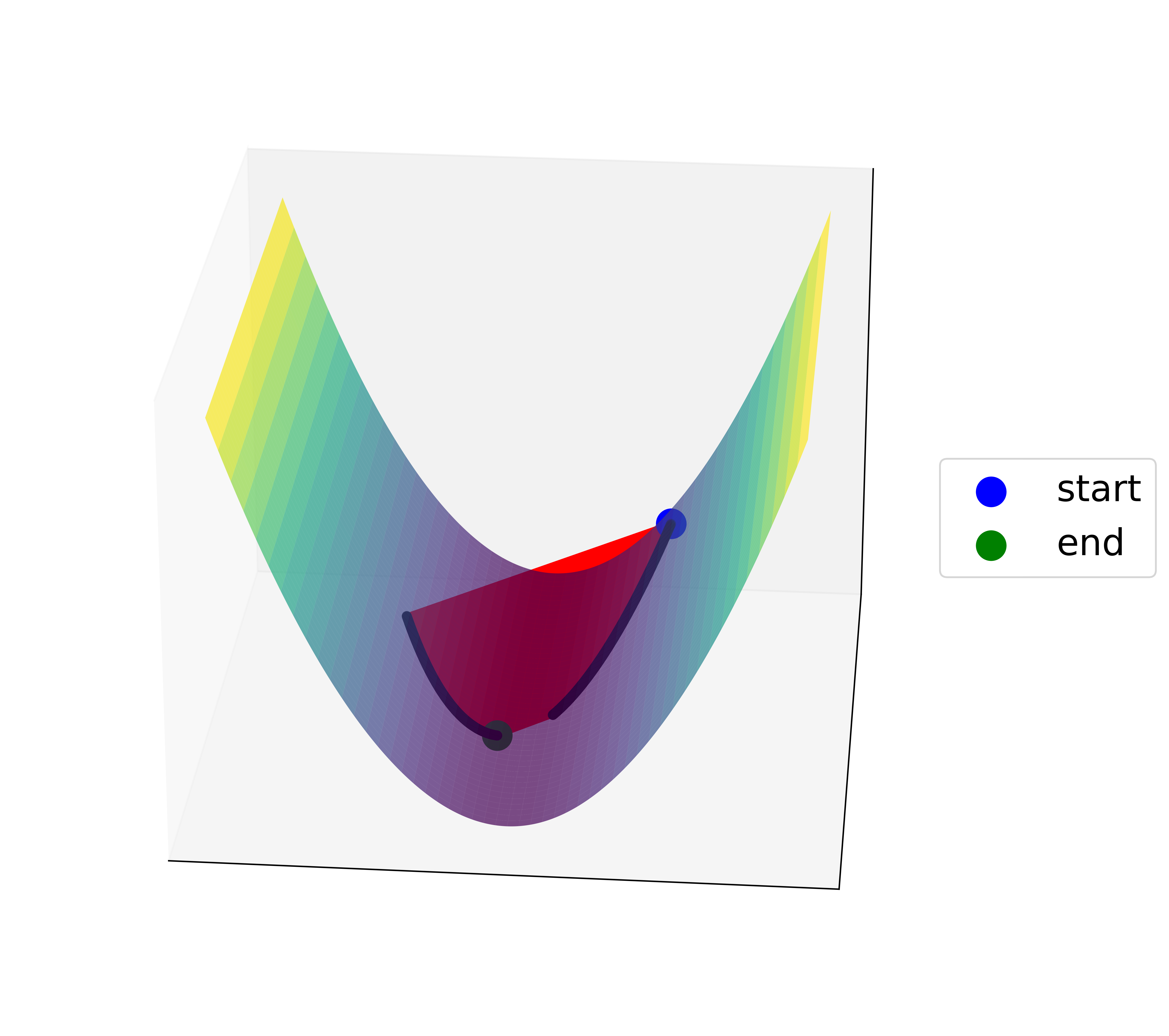}
  \caption{\textbf{Simplified EoS dynamics} on the loss landscape. Iterates oscillate stably in the normal space of the manifold of minimizers, while drifting along the tangent space in the direction of decreasing sharpness.}
  \label{fig:xy}
\end{figure}

In this section, we present an overview of key tools from bifurcation theory. For a complete exposition, we refer the reader to \citet{Kuznetsov1998}.

Let $f: \R^d \to \R^d$ be a $\mathcal{C}^3$ function that defines a discrete dynamical system
\begin{equation} \label{eq:dynamical_system}
    \vct{x}_{t+1} = f(\vct{x}_t)
\end{equation}
Denote by $\mtx{A}$, $\mtx{B}$, and $\mtx{C}$ the first, second, and third order tensor derivatives of $f$ at some point $\vct{x}_0 \in \R^d$, respectively. Bifurcation theory studies the qualitative change in behavior of the dynamical system when the Jacobian $\mtx{A}$ has so called \emph{critical} eigenvalues $\lambda \in \C$ with modulus $1$. 

Suppose that $\mtx{A}$ has $n_0$ critical eigenvalues, and let $T^c$ be the (generalized) eigenspace spanned by the corresponding (generalized) eigenvectors. A key tool in the study of bifurcations is the existence of a \emph{center manifold}.

\begin{theorem}[Center Manifold]
There is a locally-defined, $\mathcal{C}^3$, $n_0$-dimensional manifold $W^c$ satisfying the following
\begin{itemize}
    \item $W^c$ is tangent to $T^c$ at $\vct{x}_0$
    \item $W^c$ is invariant under \cref{eq:dynamical_system}
\end{itemize}
\end{theorem}

If all other eigenvalues have modulus less than $1$, then the manifold is attracting, and the \emph{reduction principle} states that the stability of the system is completely determined by the restriction of the system onto the invariant manifold $W^c$.

\textbf{Flip bifurcations.}
If $n_0 = 1$, then the bifurcation is called \emph{simple}, and the bifurcation can be classified according to the value of the critical eigenvalue $\lambda$. If $\lambda = -1$, the bifurcation is known as a flip bifurcation. At such a point, a stable periodic orbit can give rise to a new orbit with a period twice of the original. The existence of a stable period-doubled orbit is determined by the \emph{first Lyapunov coefficient},

\begin{proposition} \label{prop:first_lyapunov_coefficient}
Let $\vct{u}, \vct{v}$ the left and right eigenvectors corresponding to the simple critical eigenvalue $\lambda = -1$, respectively, with $\<\vct{u}, \vct{v}\> = 1$. The \emph{first Lyapunov coefficient} is
\begin{equation} \label{eq:first_lyapunov_coefficient}
    c_1 = \frac{1}{6} \<\vct{u}, \mtx{C}[\vct{v}]^3\> - \frac{1}{2} \<\vct{u}, \mtx{B}[\vct{v}][\vct{h}]\>
\end{equation}
where $\vct{h} = (\mtx{A} - \mtx{I})^{-1} \mtx{B}[\vct{v}]^2$. 

If $c_1 > 0$, then the bifurcation is called \emph{supercritical}, and there exists a stable period-doubled orbit near $\vct{x}_0$.
 
\end{proposition}

\textbf{Projection method.}
Proposition \ref{prop:first_lyapunov_coefficient} can be derived using the projection method. To simplify notation, assume \cref{eq:dynamical_system} has a period 1 stable orbit at $\vct{x}_0 = \vct{0}$. That is $f(\vct{0}) = \vct{0}$. Consider a Taylor expansion of $f$:
\begin{equation*}
    f(\vct{x}) = \mtx{A}\vct{x} + \frac{1}{2} \mtx{B} [\vct{x}]^2 + \frac{1}{6} \mtx{C} [\vct{x}]^3 + O(\|\vct{x}\|^4)
\end{equation*}

The two step update is
\begin{align}
    \nonumber
    f(f(\vct{x})) &= \mtx{A}^2 \vct{x} \\
    \nonumber
    &+ \frac{1}{2} \mtx{A} \mtx{B} [\vct{x}]^2 + \frac{1}{2} \mtx{B} [\mtx{A} \vct{x}]^2 \\
    \nonumber
    &+ \frac{1}{6} \mtx{A} \mtx{C} [\vct{x}]^3 + \frac{1}{2} \mtx{B}[\mtx{A}\vct{x}][\mtx{B}[\vct{x}]^2 ] + \frac{1}{6}  \mtx{C} [\mtx{A} \vct{x}]^3 \\
    \label{eq:two_step}
    &+ O(\|\vct{x}\|^4)
\end{align}

The projection method uses the left and right eigenvectors of $\mtx{A}$ to project the dynamics into the critical subspace:
\begin{lemma} \label{lemma:center_manifold_projection}
Let $\vct{u}, \vct{v}$ the left and right eigenvectors corresponding to the simple critical eigenvalue $\lambda$, respectively, with $\<\vct{u}, \vct{v}\> = 1$. The center manifold can be written as
\begin{equation} \label{eq:center_manifold_projection}
    \vct{x} = p \vct{v} + \frac{1}{2} p^2 \vct{h}_2 + \frac{1}{6} p^3 \vct{h}_3 + O(p^4)
\end{equation}
where $p = \<\vct{u}, \vct{x}\>$ is the coordinate projection onto $\vct{u}$ and $\<\vct{u}, \vct{h}_i\> = 0$ for $i = 2, 3$.

\end{lemma}

Substituting \cref{eq:center_manifold_projection} and projecting onto the left eigenvector $\vct{u}$, \cref{eq:two_step} simplifies to
\begin{align}
    \nonumber
    \<\vct{u}, f(f(\vct{x})\> &= p - p^3 \left(\frac{1}{3} \<\vct{u}, \mtx{C}[\vct{v}]^3\> + \<\vct{u}, \mtx{B}[\vct{v}][\vct{h}]]\>\right) \\
    \label{eq:two_step_center_manifold}
    &+ O(p^4)
\end{align}

where the cubic term is twice the first Lyapunov coefficient. Details are deferred to \cref{apdx:proofs}.
\section{Application to Gradient Descent} \label{sec:gd_bifurcation}

Consider a loss function $\L: \R^d \to \R$ with continuous fourth derivatives. Gradient descent with learning rate $\eta$ on $\L$ is given by the update equation
\begin{equation} \label{eq:gd}
    \vct{x}_{t+1} = \vct{x}_t - \eta \nabla \L(\vct{x}_t)
\end{equation}
This corresponds to \cref{eq:dynamical_system} with $f(\vct{x}) = \vct{x} - \eta \nabla \L(\vct{x})$. In this setting, we have $\mtx{A} = \mtx{I} - \eta \nabla^2 \L(\vct{x})$. At the edge of stability threshold, the maximum eigenvalue of the Hessian $\nabla^2 \L(\vct{x})$, which we denote by $\lambda_{\max}$, satisfies $\eta \lambda_{\max} = 2$, so $\mtx{A}$ has a critical eigenvalue $\lambda = -1$. Additionally, $\mtx{A}$ is symmetric, so the left and right eigenvectors are the same and coincide with the eigenvectors of $\nabla^2 \L(\vct{x})$.

\subsection {Simple Case} \label{sec:gd_simple_case}
First consider the case where $\L$ has an isolated minimum at $\vct{x}_*$, with $ \nabla^2 \L(\vct{x}_*) \succ 0$ and a unique top eigenvalue. Then $\L$ exhibits a simple flip bifurcation. In this setting, the first Lyapunov coefficient becomes
\begin{equation} \label{eq:first_lyapunov_coefficient_gd}
    c_1 = \frac{\eta}{2} \nabla^3 \L(\vct{x}_*)[\vct{v}_{\max}]^2[\vct{h}] - \frac{\eta}{6} \nabla^4 \L(\vct{x}_*)[\vct{v}_{\max}]^4
\end{equation}
where $\vct{v}_{\max}$ is the unit maximal eigenvector of $\nabla^2 \L(\vct{x}_*)$ and $\vct{h} = (\nabla^2 \L(\vct{x}_*))^{-1} \nabla^3 \L(\vct{x}_*)[\vct{v}_{\max}]^2$. If $c_1 > 0$, then stable period 2 oscillations exist when the learning rate $\eta$ is slightly larger than $\frac{2}{\lambda_{\max}}$, a result shown by \citet{mulayoff2026stability}. If $d = 1$ and ignoring constant factors, then the expression can be rewritten as $3 \frac{\L^{(3)}(x_*)}{\L''(x_*)} - \L^{(4)}(x_*)$, which was previously studied in \citet{chen2023edge}.

\subsection{General Case} \label{sec:gd_general_case}
In modern, overparameterized neural networks, the Hessian is almost always rank deficient \cite{sagun2018empirical, singh2021analyticinsightsstructurerank}. In fact, the loss has a manifold of minima which arise from symmetries within the network weights \cite{draxler2019essentiallybarriersneuralnetwork, simsek2021geometrylosslandscapeoverparameterized}. Thus, the simple flip bifurcation analysis is insufficient to explain the behavior of neural networks in practice.

In the general case where the Hessian may be degenerate, the zero eigenvalues of the Hessian corresponds to critical eigenvalues of $\mtx{A}$ equal to $1$, hence we no longer have a simple one-dimensional bifurcation. We make the following assumptions:
\begin{assumption} \label{assumptions:loss}
Let $\M$ be a manifold of minimizers of $\L$. Then the following hold
    \begin{itemize}
        \item \textbf{Spectral gap}. For all $\vct{x} \in \M$, $\nabla^2 \L(\vct{x})$ has a unique top eigenvalue. 

        \item \textbf{Morse-Bott}. At any $\vct{x} \in \M$ the kernel of the Hessian is equal to the tangent space of $\M$ at $\vct{x}$.
    \end{itemize}
\end{assumption}

Under the first assumption assumption, the eigenvector corresponding to the maximal eigenvalue defines the direction of maximal sharpness, and the sharpness is differentiable. A similar assumption appears in \citet{damian2023selfstabilization}. The second assumption states that the Hessian is non-degenerate in all directions normal to the manifold, and has also been used in prior analysis of neural network optimization \cite{li2022happenssgdreacheszero, arora2022understanding}.

Let $\vct{x}_* \in \M$. Consider the component of the center manifold $W_c(\vct{x}_*)$ that is normal to $\M$. This can be written as
\begin{equation} \label{eq:center_manifold_projection_normal}
    \vct{x} - \vct{x}_* = p\vct{v}_{\max} +  \frac{1}{2} p^2 \vct{h}_2 + \frac{1}{6} p^3 \vct{h}_3 + O(p^4)
\end{equation}

Following the projection method as in \cref{eq:two_step_center_manifold}, we can derive the following: 
\begin{lemma} \label{lemma:two_step_stability_normal}
Let $\vct{x}_* \in \M$. Set $c_1(\vct{x}_*)$ as in \cref{eq:first_lyapunov_coefficient_gd}, except that $\vct{h} = (\nabla^2 \L(\vct{x}_*))^{\dag} \nabla^3 \L(\vct{x}_*)[\vct{v}_{\max}]^2$. Let $\Pi_{N_{\vct{x}_*} \M}$ be the projection into the normal space of $\M$ at $\vct{x}_*$. Then there exists $\vct{x}_s$ satisfying \cref{eq:center_manifold_projection_normal} and
\begin{equation*}
    \Pi_{N_{\vct{x}_*} \M} (f(f(\vct{x}_s)) - \vct{x}_*) = \Pi_{N_{\vct{x}_*} \M} (\vct{x}_s - \vct{x}_*)
\end{equation*}
and $\vct{x}_s$ is stable equilibrium for the projected dynamics in $N_{\vct{x}_*} \M$.
\end{lemma}

Compared to Proposition \ref{prop:first_lyapunov_coefficient}, $\vct{h}_2$ is now defined using the pseudoinverse, and the two-step fixed point only appears when projecting into the normal space of the manifold of minimizers. 

In order to fully understand the evolution of the system, we must also understand the dynamics in the tangent space of $\M$. Indeed, using the projection method again gives the following lemma:

\begin{lemma} \label{lemma:two_step_drift_tangent}
Let $\vct{x}_* \in \M$, and $\vct{x} \in W^c$ following \cref{eq:center_manifold_projection_normal}. Let $\Pi_{T_{\vct{x}_*} \M}$ be the projection onto the tangent space of $\M$ at $\vct{x}_*$. Then
\begin{align} 
    \nonumber
    \Pi_{T_{\vct{x}_*} \M} (f(f(\vct{x})) - \vct{x}_*) &= - \eta p^2 \Pi_{T_{\vct{x}_*} \M} \nabla^3 \L(\vct{x}_*)[\vct{v}_{\max}]^2 \\
    \label{eq:two_step_drift_tangent}
    &+ O(\eta p^3)
\end{align}
\end{lemma}

Note that $\nabla^3 \L(\vct{x}_*)[\vct{v}_{\max}]^2$, is the gradient of the sharpness. So the previous Lemma shows that the iterates drift along the manifold of minimizers in the direction of decreasing sharpness.

We now have a complete picture of the training dynamics of gradient descent on the edge of stability based on the tangent and normal space decomposition. In the normal subspace of $\M$, iterates approach a stable period two orbit of the projected dynamics in the normal space. Meanwhile, in the tangent space, the iterates drift in the direction of decreasing sharpness. As long as the first Lyapunov coefficient $c_1$ stays positive along the manifold, the iterates can remain stable. And if the sharpness eventually decays to below the edge of stability threshold, the iterates can in fact converge to the minimizing manifold. We formalize the results in the following theorem:

\begin{theorem} \label{thm:main_result}
Suppose $\L: \R^d \to \R$ has continuous fourth derivatives and the manifold of minima $\M$ satisfies Assumption \ref{assumptions:loss}. Suppose that for  $\vct{x}_* \in \M$, the following hold
\begin{enumerate}
    \item $c_1(\vct{x}_*) > 0$
    \item $\Pi_{T_{\vct{x}_*} \M} \nabla^3 \L(\vct{x}_*)[\vct{v}_{\max}]^2 \neq 0$
\end{enumerate}
Then there exists a neighborhood $N$ of $\vct{x}_*$ such that gradient descent with initialization $\vct{x}_0 \in N$ and learning rate $\eta = \frac{2}{\lambda_{\max}(\vct{x}_*)}$ converges to a point on $\M$.
\end{theorem}

This theorem defines the rigorous foundations for the convergence on the edge of stability. Whereas classical results via the descent lemma require $\eta < \frac{2}{\lambda_{\max}}$, \cref{thm:main_result} applies precisely on the EoS threshold $\eta = \frac{2}{\lambda_{\max}}$. Note that Condition 2 guarantees that the neighborhood $N$ contains points with sharpness both above and below the EoS threshold.

The assumptions of \cref{thm:main_result} are general and can be \emph{applied directly to real-world deep neural networks}. Preliminary work has already shown that fully-connected multi-layer perceptrons satisfy Condition 1 \cite{gan2026product}.
\section{Illustrative Example} \label{sec:discussion}

A prior line of work on edge of stability training dynamics focused on minimalist, two-dimensional losses of the form $\L(x,y) = f(xy)$. Suppose $f$ has an isolated minimum at $z_*$ with $f''(z_*) > 0$. Then $\L$ has a one dimensional manifold of minima along the hyperbola $xy = z_*$.

\citet{gan2026product} unified this line of work by showing that all non-quadratic losses of the previous form for which EoS convergence has been proven satisfy a condition which they called \emph{product-stability}, and that this condition is sufficient for convergence.

\begin{definition} [\citet{gan2026product}]
    Let $f: \R \to \R$ have continuous fourth derivatives. The \emph{product-stability} of $f$ at $z \in \R$ is defined as
    \begin{equation} \label{eq:def_product_stability}
        \alpha_f(z) = 3( f^{(3)}(z))^2 - f^{(4)} (z) f''(z)
    \end{equation}
    If $\alpha_f(z) > 0$, then we say $f$ is \emph{product-stable} at $z$.
\end{definition}

Here, we will show that this setting is a special case of the bifurcation analysis in \cref{sec:gd_bifurcation}. The first clue to this connection is that $\alpha_f$ is, up to scaling, the first Lyapunov coefficient for the scalar function $f$. But how does this relate to the first Lyapunov coefficient of the two-dimensional loss $\L$? The following Lemma shows that they are intrinsically related.

\begin{lemma} \label{lemma:product_stability_lyapunov_coefficient}
Let $f: \R \to \R$ have continuous fourth derivatives and consider $\L(x, y) = f(xy)$. The first Lyapunov coefficient of $\L$ is
\begin{align*}
    c_1 &= \frac{\eta}{6} \bigg( \left( 3 \frac{f^{(3)}(xy)}{f''(xy)} - f^{(4)}(xy) \right)(x^2 + y^2)^2 \\
    &+ 24 f^{(3)}(xy) xy + 96f''(xy) \frac{x^2 y^2}{(x^2 + y^2)} \bigg)
\end{align*}
\end{lemma}

As we move along the hyperbola $xy = z_*$, when $x^2 + y^2$ is large, the first term dominates, and therefore $c_1$ has the same sign as $\alpha_f$.

The following lemma computes the gradient of the sharpness:
\begin{lemma} \label{lemma:product_stability_sharpness_gradient}
Let $\L$ be as in Lemma \ref{lemma:product_stability_lyapunov_coefficient}. Suppose $x y = z_*$ with $f'(z_*) = 0$. Then
\begin{align*}
    \nabla^3 \L(x, y)[\vct{v}_{\max}]^2 &= f^{(3)}(x y) (x^2 + y^2) \begin{pmatrix}
        y \\ x
    \end{pmatrix} \\
    &+ 2 \frac{f''(x y)}{x^2 + y^2} \begin{pmatrix}
        x(x^2 + 2y^2) \\ y(2x^2 + y^2)
    \end{pmatrix}
\end{align*}
Let $\Pi_\M$ be the projection onto the tangent space of the manifold $xy = z_*$. Then
\begin{align*}
    \Pi_\M \nabla^3 \L(x, y)[\vct{v}_{\max}]^2 &= 2 f''(x y) \frac{x^2 - y^2}{x^2 + y^2} \begin{pmatrix}
        x \\ -y
    \end{pmatrix}
\end{align*}
\end{lemma}

Since $f''(z_*) > 0$, $\Pi_\M \nabla^3 \L(x, y)[\vct{v}_{\max}]^2$ is nonzero as long as $x^2 - y^2$ is nonzero.

Thus, the previous two lemmas establish that the product stability construction from \citet{gan2026product} is an illustrative example of \cref{thm:main_result}.

\section{Conclusion}

This work uses bifurcation theory to define a framework for explaining the stable convergence of gradient descent on the edge of stability. By decomposing the dynamics into the tangent and normal components with respect to the manifold of minimizers, we derive predictable dynamics in each subspace that lead to convergence. The framework generalizes previous constructions in the EoS literature and can be applied directly to modern, overparameterized neural networks. 

There are still aspects to EoS training which remain open. While empirical evidence suggests that minima of practical machine learning models have positive first Lyapunov coefficient, there is no theoretical framework which can explain why this is the case from first principles. Additionally, the underlying mechanism behind progressive sharpening, through which training iterates enter the EoS regime, remains unresolved.

\section*{Impact Statement}

This paper presents work whose goal is to advance the field of Machine
Learning. There are many potential societal consequences of our work, none
which we feel must be specifically highlighted here.

\bibliography{reference}
\bibliographystyle{icml2026}


\newpage
\appendix
\onecolumn

\section{Proofs} \label{apdx:proofs}

\subsection{Preliminaries}

We write $\< \cdot, \cdot \>$ for the standard Euclidean inner product and $\| \cdot \|$ for the induced norm. For a smooth function $\L: \R^d \to |R$, we denote its gradient by $\nabla \L$, its Hessian by $\nabla^2 \L$, and its
k-th order derivative tensor by $\nabla^k \L$. For a symmetric bilinear (or multilinear) form $\mtx{B}$, we write $\mtx{B}[\vct{u}][\vct{v}]$ for the application to vectors $\vct{u}, \vct{v}$, and $\mtx{B}[\vct{v}]^2$ as shorthand for $\mtx{B}[\vct{v}][\vct{v}]$. Similarly $\mtx{C}[\vct{v}]^3 = \mtx{C}[\vct{v}][\vct{v}][\vct{v}]$ for a trilinear form $\mtx{C}$. 

The pseudoinverse of a matrix $\mtx{M}$ is denoted $\mtx{M}^{\dag}$. For a submanifold $\mathcal{M} \subset \R^d$ and a point $\vct{x} \in \mathcal{M}$, we write $T_x \mathcal{M}$ and $N_{\vct{x}}\mathcal{M}$ for the tangent and normal spaces of $\M$ at $\vct{x}$, respectively, and $\Pi_{T_{\vct{x}}\mathcal{M}}$, $\Pi_{N_{\vct{x}}\mathcal{M}}$ for the corresponding orthogonal projections. 

The largest eigenvalue of a symmetric matrix $\mtx{H}$ is denoted $\lambda_{\max}(\mtx{H})$, and $\vct{v}_{\max}$ denotes a corresponding unit eigenvector. We refer to $\lambda_{\max}(\nabla^2 \L(\vct{x}))$ as the sharpness at $\vct{x}$. The learning rate is denoted $\eta > 0$, and the gradient descent map is $f(\vct{x}) = \vct{x} - \eta \nabla \L(\vct{x})$. 

The center manifold of $f$ at a point $\vct{x}$ is denoted $W^c(x_0)$. The first Lyapunov coefficient associated to a flip bifurcation is denoted $c_1$.


\subsection{Derivation of \cref{eq:two_step_center_manifold}} \label{apdx:two_step_center_manifold}
Substituting $\vct{x} = p \vct{v} + \frac{1}{2} p^2 \vct{h}_2 + \frac{1}{6} p^3 \vct{h}_3 + O(p^4)$, we have
\begin{align*}
    \<\vct{u}, f(f(\vct{x}))\> &= \vct{u}^{\top} \mtx{A}^2 (p \vct{v} + \frac{1}{2} p^2 \vct{h}_2 + \frac{1}{6} p^3 \vct{h}_3) \\
    &+ \frac{1}{2} \vct{u}^{\top} \mtx{A} \mtx{B} [p \vct{v} + \frac{1}{2} p^2 \vct{h}_2]^2 + \frac{1}{2} \vct{u}^{\top} \mtx{B} [\mtx{A} (p \vct{v} + \frac{1}{2} p^2 \vct{h}_2)]^2 \\
    &+ \frac{1}{6} \vct{u}^{\top} \mtx{A} \mtx{C} [\vct{v}]^3 + \frac{1}{2} \vct{u}^{\top} \mtx{B}[p\mtx{A}\vct{v}][\mtx{B}[p\vct{v}]^2 ] + \frac{1}{6} \vct{u}^{\top} \mtx{C} [p \mtx{A} \vct{v}]^3 \\
    &+ O(p^4) \\
\end{align*}

Using the fact that $\vct{u}$ and $\vct{v}$ are left and right eigenvectors of $\mtx{A}$ with eigenvalue $-1$, we have 
\begin{align*}
    \<\vct{u}, f(f(\vct{x}))\> &= \vct{u}^{\top} (p \vct{v} + \frac{1}{2} p^2 \vct{h}_2 + \frac{1}{6} p^3 \vct{h}_3) \\
    &-\frac{1}{2} \vct{u}^{\top} \mtx{B} [p \vct{v} + \frac{1}{2} p^2 \vct{h}_2]^2 + \frac{1}{2} \vct{u}^{\top} \mtx{B} [- p \vct{v} + \frac{1}{2} p^2  \mtx{A} \vct{h}_2]^2 \\
    &-\frac{1}{6} p^3 \vct{u}^{\top} \mtx{C} [\vct{v}]^3 - \frac{1}{2} p^3 \vct{u}^{\top} \mtx{B}[\vct{v}][\mtx{B}[\vct{v}]^2 ] - \frac{1}{6} p^3 \vct{u}^{\top} \mtx{C} [\vct{v}]^3 \\
    &+ O(p^4) \\
\end{align*}

Now expanding and using $\<\vct{u}, \vct{v}\> = 1, \<\vct{u}, \vct{h}_i\> = 0$, we have
\begin{align*}
    \<\vct{u}, f(f(\vct{x}))\> &= p  \\
    &-\frac{1}{2} p^2 \vct{u}^{\top} \mtx{B} [\vct{v}]^2 - \frac{1}{2} p^3 \vct{u}^{\top} \mtx{B}[\vct{v}][\vct{h}_2] + \frac{1}{2} p^2 \vct{u}^{\top} \mtx{B} [\vct{v}]^2 - \frac{1}{2} p^3 \vct{u}^{\top} \mtx{B}[\vct{v}][\vct{A}\vct{h}_2] \\
    &-\frac{1}{3} p^3 \vct{u}^{\top} \mtx{C} [\vct{v}]^3 - \frac{1}{2} p^3 \vct{u}^{\top} \mtx{B}[\vct{v}][\mtx{B}[\vct{v}]^2 ] \\
    &+ O(p^4) \\
    &= p -\frac{1}{3} p^3 \vct{u}^{\top} \mtx{C} [\vct{v}]^3 - \frac{1}{2}p^3\vct{u}^{\top} \mtx{B}[\vct{v}]\left[(\mtx{I} + \mtx{A}) \vct{h}_2 + \mtx{B}[\vct{v}]^2\right] + O(p^4)
\end{align*}
From here, we define the critical constant as half the coefficient of the cubic term
\begin{equation*}
    c_1 = \frac{1}{6} \vct{u}^{\top} \mtx{C} [\vct{v}]^3 + \frac{1}{4} \vct{u}^{\top} \mtx{B}[\vct{v}]\left[(\mtx{I} + \mtx{A}) \vct{h}_2 + \mtx{B}[\vct{v}]^2\right]
\end{equation*}
If $c_1 > 0$, then this is a supercritical flip bifurcation, and there exists $\vct{x}$ such that $\<\vct{u}, f(f(\vct{x}))\> = \<\vct{u}, \vct{x}\>$. By the reduction principle, this implies that 
\begin{equation} \label{eq:two_step_stability_def}
    f(f(\vct{x})) = \vct{x}    
\end{equation}
Using this fact, we can compare the $p^2$ term in \cref{eq:two_step_stability_def} to get
\begin{align*}
    \vct{h}_2 &= \mtx{A}^2 \vct{h}_2 + \mtx{A} \mtx{B}[\vct{v}]^2 + \mtx{B}[\vct{v}]^2 
\end{align*}
Hence we have $(\mtx{I} - \mtx{A}^2) \vct{h}_2 = (\mtx{I} + \mtx{A}) \mtx{B}[\vct{v}]^2$. Since $\ker(\mtx{I} + \mtx{A})$ is spanned by $\vct{v}$, it follows that
\begin{equation*}
    (\mtx{I} - \mtx{A}) \vct{h}_2 = \mtx{B}[\vct{v}]^2 + \alpha \vct{v}
\end{equation*}
Now $\mtx{I} - \mtx{A}$ is invertible since $\mtx{A}$ has no eigenvalues equal to 1, hence
\begin{equation*}
    \vct{h}_2 = (\mtx{I} - \mtx{A})^{-1} \mtx{B}[\vct{v}]^2 + \frac{1}{2}\alpha \vct{v}
\end{equation*}
Set $\vct{h} = (\mtx{I} - \mtx{A})^{-1} \mtx{B}[\vct{v}]^2$. Then $(\mtx{I} + \mtx{A}) \vct{h} = (\mtx{I} + \mtx{A}) \vct{h}_2$, and we can simplify

\begin{align*}
    c_1 &= \frac{1}{6} \vct{u}^{\top} \mtx{C} [\vct{v}]^3 + \frac{1}{4} \vct{u}^{\top} \mtx{B}[\vct{v}]\left[(\mtx{I} + \mtx{A}) \vct{h} + (\mtx{I} - \mtx{A}) \vct{h} \right] \\
    &= \frac{1}{6} \vct{u}^{\top} \mtx{C} [\vct{v}]^3 + \frac{1}{2} \vct{u}^{\top} \mtx{B}[\vct{v}][\vct{h}]
\end{align*}
as desired.

\subsection{Proof of Lemma \ref{lemma:two_step_stability_normal}} \label{apdx:two_step_stability_normal}
Notice that \cref{eq:center_manifold_projection_normal} has the same form as in \cref{apdx:two_step_center_manifold}, so we can follow the steps. The only difference is that $\mtx{I} - \mtx{A}$ is no longer invertible. However, this is not a problem as $\vct{h}_2$ is orthogonal to the center manifold, hence orthogonal to $\ker(\mtx{I} - \mtx{A})$. Thus the calculation works with the inverse replaced with the pseudoinverse.

\subsection{Proof of Lemma \ref{lemma:two_step_drift_tangent}}
From \cref{eq:two_step}, we have
\begin{align*}
   \Pi_{T_{\vct{x}_*} \M} (f(f(\vct{x})) - \vct{x}_*)  &= \Pi_{T_{\vct{x}_*} \M} \mtx{A}^2 (p \vct{v} + \frac{1}{2} p^2 \vct{h}_2) \\
    &+ \frac{1}{2} \Pi_{T_{\vct{x}_*} \M} \mtx{A} \mtx{B} [p \vct{v} + \frac{1}{2} p^2 \vct{h}_2]^2 + \frac{1}{2} \Pi_{T_{\vct{x}_*} \M} \mtx{B} [\mtx{A} (p \vct{v} + \frac{1}{2} p^2 \vct{h}_2)]^2 \\
    &+ O(p^3) \\
    &= \Pi_{T_{\vct{x}_*} \M} \mtx{A}^2 (p \vct{v} + \frac{1}{2} p^2 \vct{h}_2) \\
    &+ \frac{1}{2} p^2 \Pi_{T_{\vct{x}_*} \M} \mtx{A} \mtx{B} [\vct{v}]^2 + \frac{1}{2} p^2 \Pi_{T_{\vct{x}_*} \M} \mtx{B} [\mtx{A} \vct{v}]^2 \\
    &+ O(p^3)
\end{align*}
But $\Pi_{T_{\vct{x}_*} \M} \mtx{A} = 0$, hence
\begin{align*}
   \Pi_{T_{\vct{x}_*} \M} (f(f(\vct{x})) - \vct{x}_*)  &= \frac{1}{2} p^2 \Pi_{T_{\vct{x}_*} \M} \mtx{B} [\mtx{A} \vct{v}]^2 + O(p^3)
\end{align*}
Finally, using the fact that $\vct{v}$ is an eigenvector of $\mtx{A}$,
\begin{align*}
   \Pi_{T_{\vct{x}_*} \M} (f(f(\vct{x})) - \vct{x}_*)  &= \frac{1}{2} p^2 \Pi_{T_{\vct{x}_*} \M} \mtx{B} [\vct{v}]^2 + O(p^3)
\end{align*}
Now $\mtx{B} = - \eta \nabla^3 \L$, giving the desired result.

\subsection{Proof of \cref{thm:main_result}}
By the reduction principle, it suffices to consider the behavior on the center manifold. Define $M: N \to \M$ such that $M(\vct{x})$ is the orthogonal projection of $\vct{x}$ onto the manifold $\M$. Then
\begin{align*}
    \nabla_{\vct{x}} \lambda_{\max}(M(\vct{x})) &= (\mtx{J}_M(\vct{x}))^{\top} (\nabla \lambda_{\max}) (M(\vct{x})) \\
    &= \Pi_{T_{M(\vct{x})} \M} (\mtx{I} - p \mtx{S})^{-1}  \nabla^3 \L(M(\vct{x}))[\vct{v}_{\max}]^2
\end{align*}
where $\mtx{S}$ is the shape operator, $\vct{v}_{\max}$ is the top eigenvector of $\nabla^2 \L(M(\vct{x}))$, and $p = \<\vct{v}_{\max}, \vct{x} - M(\vct{x})\>$. Then using Lemma \ref{lemma:two_step_drift_tangent}
\begin{align*}
    \lambda_{\max}(M(f(f(\vct{x})))) &=  \lambda_{\max}(M(\vct{x})) + \<f(f(\vct{x})) - \vct{x}, \nabla_{\vct{x}} \lambda_{\max}(M(\vct{x}))\> + O(\|\Pi_{T_{\vct{x}_*} \M} (f(f(\vct{x})) - \vct{x})\|^2) \\
    &= \lambda_{\max}(M(\vct{x})) + \<\Pi_{T_{M(\vct{x})} \M} f(f(\vct{x})) - \vct{x}, (\mtx{I} - p \mtx{S})^{-1} \vct{z}\>  + O(\|\Pi_{T_{\vct{x}_*} \M} (f(f(\vct{x})) - \vct{x})\|^2) \\
    &= \lambda_{\max}(M(\vct{x})) - \frac{\eta}{2} p^2 \vct{z}^{\top} \Pi_{T_{M(\vct{x})} \M} (\mtx{I} - p \mtx{S})^{-1} \vct{z} + O(p^3)
\end{align*}
where $\vct{z} = \nabla^3 \L(M(\vct{x}))[\vct{v}_{\max}]^2$. Condition 2 guarantees that $\vct{z}^{\top} \Pi_{T_{M(\vct{x})} \M} (\mtx{I} - p \mtx{S})^{-1} \vct{z}$ is strictly negative, so $\lambda_{\max}(M(\vct{x}))$ is decreasing with respect to the two step iterates in a neighborhood of $\vct{x}_*$. Moreover, this sequence is nonnegative and only stabilizes for $\vct{x} \in \M$, hence we obtain convergence in $\M$ if we can show that the iterates remain in a neighborhood where the sequence is decreasing.

Indeed, by Condition 1 and Lemma \ref{lemma:two_step_drift_tangent}, there exists stable points in the normal space dynamics. Since these points are attractive, we have
\begin{equation*}
    \|\Pi_{N_{\vct{x}_*} \M} (f(f(\vct{x})) - \vct{x}_s)\| \leq \|\Pi_{N_{\vct{x}_*} \M} (\vct{x} - \vct{x}_s)\|
\end{equation*}
This guarantees that the iterates do not diverge away from the neighborhood, completing the proof.

\subsection{Proof of Lemma \ref{lemma:product_stability_lyapunov_coefficient}} \label{apdx:lemma:product_stability_lyapunov_coefficient}
Compute
\begin{align*}
    \partial_x \L &= f'(xy) y \\
    \partial_y \L &= f'(xy) x \\
    \partial_{xx} \L &= f''(xy) y^2 \\
    \partial_{xy} \L &= f''(xy) xy + f'(xy) \\
    \partial_{yy} \L &= f''(xy) x^2 \\
    \partial_{xxx} \L &= f^{(3)}(xy) y^3 \\
    \partial_{xxy} \L &= f^{(3)}(xy) xy^2 + f''(xy) y \\
    \partial_{xyy} \L &= f^{(3)}(xy) x^2 y + f''(xy) x \\
    \partial_{yyy} \L &= f^{(3)}(xy) x^3 \\
\end{align*}
When $f'(xy) = 0$, we have
\begin{align*}
    \nabla^2 \L(x, y) &= f''(xy) \begin{pmatrix}
        y^2 & xy \\
        xy & x^2 
    \end{pmatrix}
\end{align*}
Hence 
\begin{equation*}
    \vct{v}_{\max} = \frac{1}{\sqrt{x^2 + y^2}} \begin{pmatrix}
    y \\ x
\end{pmatrix}
\end{equation*}

Let $\vct{u} = \nabla^3 \L(x, y)[\vct{v}_{\max}]^2$. We now compute 
\begin{align} \label{eq:product_stability_sharpness_gradient}
    \vct{u} &= \begin{pmatrix}
        f^{(3)}(xy) y (x^2 + y^2) + 2 f''(xy) x \frac{x^2 + 2y^2}{x^2 + y^2} \\
        f^{(3)}(xy) x (x^2 + y^2) + 2 f''(xy) y \frac{2x^2 + y^2}{x^2 + y^2}
    \end{pmatrix}
\end{align}

 Then
\begin{align*}
    \nabla^3 \L(x, y)[\vct{v}_{\max}]^2 [\vct{h}] &= \vct{u}^{\top} (\nabla^2 \L(xy))^{\dag} \vct{u} \\
    &= \frac{1}{f''(xy)(x^2 + y^2)} (\vct{u}^{\top} \vct{v}_{\max})^2 \\
    &= \frac{(f^{(3)}(xy))^2}{f''(xy)} (x^2 + y^2)^2 + 12 f^{(3)}(xy) xy + 36 f''(xy) \frac{x^2 y^2}{(x^2 + y^2)^2}
\end{align*}

On the other hand
\begin{align*}
    \nabla^4 \L(xy) [\vct{v}_{\max}]^4 &= f^{(4)}(xy) (x^2 + y^2)^2 + 12 f^{(3)}(xy) xy + f''(xy) \frac{x^2 y^2}{(x^2 + y^2)^2}
\end{align*}

Substituting the above two expressions gives the desired result.

\subsection{Proof of Lemma \ref{lemma:product_stability_sharpness_gradient}}
Observe that $\M$ is spanned by
\begin{equation*}
    \vct{w} = \frac{1}{\sqrt{x^2 + y^2}} \begin{pmatrix}
    x \\ -y
\end{pmatrix}
\end{equation*}

Using \cref{eq:product_stability_sharpness_gradient}, we have
\begin{align*}
    \Pi_\M \nabla^3 \L(x, y)[\vct{v}_{\max}]^2 [\vct{h}] &= \<\vct{u}, \vct{w}\> \vct{w} \\
    &= 2 f''(x y) \frac{x^2 - y^2}{x^2 + y^2} \begin{pmatrix}
        x \\ -y
    \end{pmatrix}
\end{align*}

This completes the proof.

\end{document}